\documentclass[conference]{IEEEtran}
\IEEEoverridecommandlockouts
\usepackage{cite}
\usepackage{amsmath,amssymb,amsfonts}
\usepackage{algorithmic}
\usepackage{graphicx}
\usepackage{textcomp}
\usepackage{xcolor}

\usepackage{graphicx}
\usepackage{booktabs}
\usepackage{subcaption}
\usepackage{url}
\usepackage{wrapfig}
\usepackage{soul}
\usepackage{amsmath,amssymb}
\usepackage[linesnumbered,lined,commentsnumbered]{algorithm2e}

\def\BibTeX{{\rm B\kern-.05em{\sc i\kern-.025em b}\kern-.08em
    T\kern-.1667em\lower.7ex\hbox{E}\kern-.125emX}}
\begin{document}

\title{Semantic Reasoning from Model-Agnostic Explanations\\

\thanks{Slovenian  Research  Agency}
}

\author{\IEEEauthorblockN{Timen Stepi\v{s}nik Perdih}
\IEEEauthorblockA{
\textit{Jo\v{z}ef Stefan Institute}\\
Ljubljana, Slovenia \\
tstepisnikp@gmail.com}
\and
\IEEEauthorblockN{Nada Lavra\v{c}}
\IEEEauthorblockA{
\textit{Jo\v{z}ef Stefan Institute}\\
Ljubljana, Slovenia\\
\textit{University of Nova Gorica} \\
Nova Gorica, Slovenia}
\and
\IEEEauthorblockN{Bla\v{z} \v{S}krlj}
\IEEEauthorblockA{
\textit{Jo\v{z}ef Stefan International Postgraduate School}\\
\textit{Jo\v{z}ef Stefan Institute}, \\
Ljubljana, Slovenia \\
blaz.skrlj@ijs.si}
}

\maketitle
\begin{abstract}
With the wide adoption of black-box models, instance-based \emph{post hoc} explanation tools, such as LIME and SHAP became increasingly popular.
These tools produce explanations, pinpointing contributions of key features associated with a given prediction. However, the obtained explanations remain at the raw feature level and are not necessarily understandable by a human expert without extensive domain knowledge. We propose ReEx (Reasoning with Explanations), a method applicable to explanations generated by arbitrary instance-level explainers, such as SHAP. By using background knowledge in the form of ontologies, ReEx generalizes instance explanations in a least general generalization-like manner. The resulting symbolic descriptions are specific for individual classes and offer generalizations based on the explainer's output. The derived semantic explanations are potentially more informative, as they describe the key attributes in the context of more general background knowledge, e.g., at the biological process level. We showcase ReEx's performance on nine biological data sets, showing that compact, semantic explanations can be obtained and are more informative than generic ontology mappings that link terms directly to feature names. ReEx is offered as a simple-to-use Python library and is compatible with tools such as SHAP and similar. To our knowledge, this is one of the first methods that directly couples semantic reasoning with contemporary model explanation methods. This paper is a preprint. Full version's doi is: 10.1109/SAMI50585.2021.9378668
\end{abstract}
\begin{IEEEkeywords}
model explanations, reasoning, generalization, SHAP, machine learning, explainable AI
\end{IEEEkeywords}

\section{Introduction}
\IEEEPARstart{T}{here} is a growing demand for machine learning approaches that are not only well-performing but also \emph{trustworthy}, transparent and \emph{explainable} \cite{DBLP:journals/corr/abs-1712-09923}.
Methods like LIME \cite{ribeiro2016should} and SHAP \cite{NIPS2017_7062} have been proposed to extract the contributions of the most important features to explain a given model's prediction, but these features are often still not fully understandable by the user (e.g., gene symbols). This paper presents a way of reasoning from model explanations called \textbf{ReEx} (Reasoning from Explanations), a method that generalizes these features in the context of a given collection of \emph{background knowledge} into explanations comprised of more understandable (semantic) terms.
This paper discusses the following advances over the state of the art: 
\begin{enumerate}
    \item We propose ReEx, a method capable of linking background knowledge in the form of ontologies to explanations generated by widely accepted approaches such as SHAP. ReEx performs efficient term generalization whilst considering an arbitrary number of relations (between the terms), \emph{automatically}.
    \item The added value of ReEx with respect to its performance is demonstrated on nine real-life gene expression data sets linked to \emph{Gene Ontology} \cite{gene2019gene}, where compact, \emph{interpretable} explanations consisting of semantic terms are obtained as a result of generalization from the attributes, recognized as relevant by SHAP~\cite{NIPS2017_7062}.
    \item We propose an information-theoretic measure of generalization derived from the information content of individual terms (or sets of terms) we refer to as \textsc{GenQ}. The measure is used to directly assess the generalization performance and is arguably more suitable for this task than the commonly employed information content.
    \item The proposed solution is presented as a simple-to-use Python package, useful off-the-shelf alongside the existing model explainers.
\end{enumerate}

\section{Related work}
\label{sec:related}
In this section, we present the related work, ranging from the notion of model explanation and semantic data mining to the more recent approaches which attempt to join the two sub-fields by treating semantic background knowledge as \emph{graph-theoretic objects}, directly useful for applications in machine learning.

General solutions for explanations based on individual instances were proposed more than a decade ago \cite{robnik2008explaining}, offering promising solutions to better understand classifiers such as the support vector machines and similar.
Explanation of black-box models, however, has resurfaced in the recent years, along with the revival of neural networks and the development of deep neural networks \cite{goodfellow2016deep}. Such multi-million parameter neural networks are able to associate beyond the capabilities of any other learners when considering inputs such as images, texts and more recently - graphs \cite{yang2015network}. Methods, such as LIME \cite{ribeiro2016should}, SHAP \cite{NIPS2017_7062} and similar \cite{biran2017explanation,strumbelj2014}, were introduced and have, by offering simple-to-use APIs, become widely used throughout the industry and science. For example, a recent review~\cite{biran2017explanation} elaborates on the importance of understanding the causality of the learned representations via explanations. It explains the difference between \emph{post-hoc} systems, which aim to provide the local explanations for a specific decision, making
it reproducible on demand instead of explaining the whole systems' behavior (LIME), and \emph{ante-hoc} systems, which are interpretable by design and are considered as the \emph{white-box approaches} (linear regression, decision trees, rules).
Further, the recent work on the use of machine learning (ML) approaches \cite{DBLP:journals/corr/abs-1710-00794} argues that blindly accepting the outcome of an ML system is a dangerous practice, which is adopted by many \emph{out of necessity}, or by choice. To overcome this a ML system should provide an explanation of its decision-making process.

The use of background knowledge for improving the understandability of machine learning systems has also been referred to as \emph{semantic data mining}\cite{dou2015semantic}. For example, in relational domains, the CBSSD methodology \cite{Skrlj2019} focuses on data mining tasks that use ontologies as background knowledge when explaining emergent structures in complex networks. Similarly, the NetSDM approach \cite{JMLR:v20:17-066} explored how redundant ontologies are for the purposes of inductive rule learning. Furthermore, promising results were observed when attempting to account for explicit semantics during image classification \cite{dou2015semantic}, and using inductive logic programming to obtain relational explanations (of image classifications) \cite{alephXAI}.
The ReEx approach presented in the following sections attempts to bridge the gap between the plethora of available relational background knowledge sources and explainable models, offering a fast and simple to use method, complementary to the existing model explainers incapable of exploiting existing background knowledge.

\section{ReEx - Reasoning from model explanations}
\label{sec:reex}
\begin{figure}[t!]
    \centering
    \includegraphics[width = .99\linewidth]{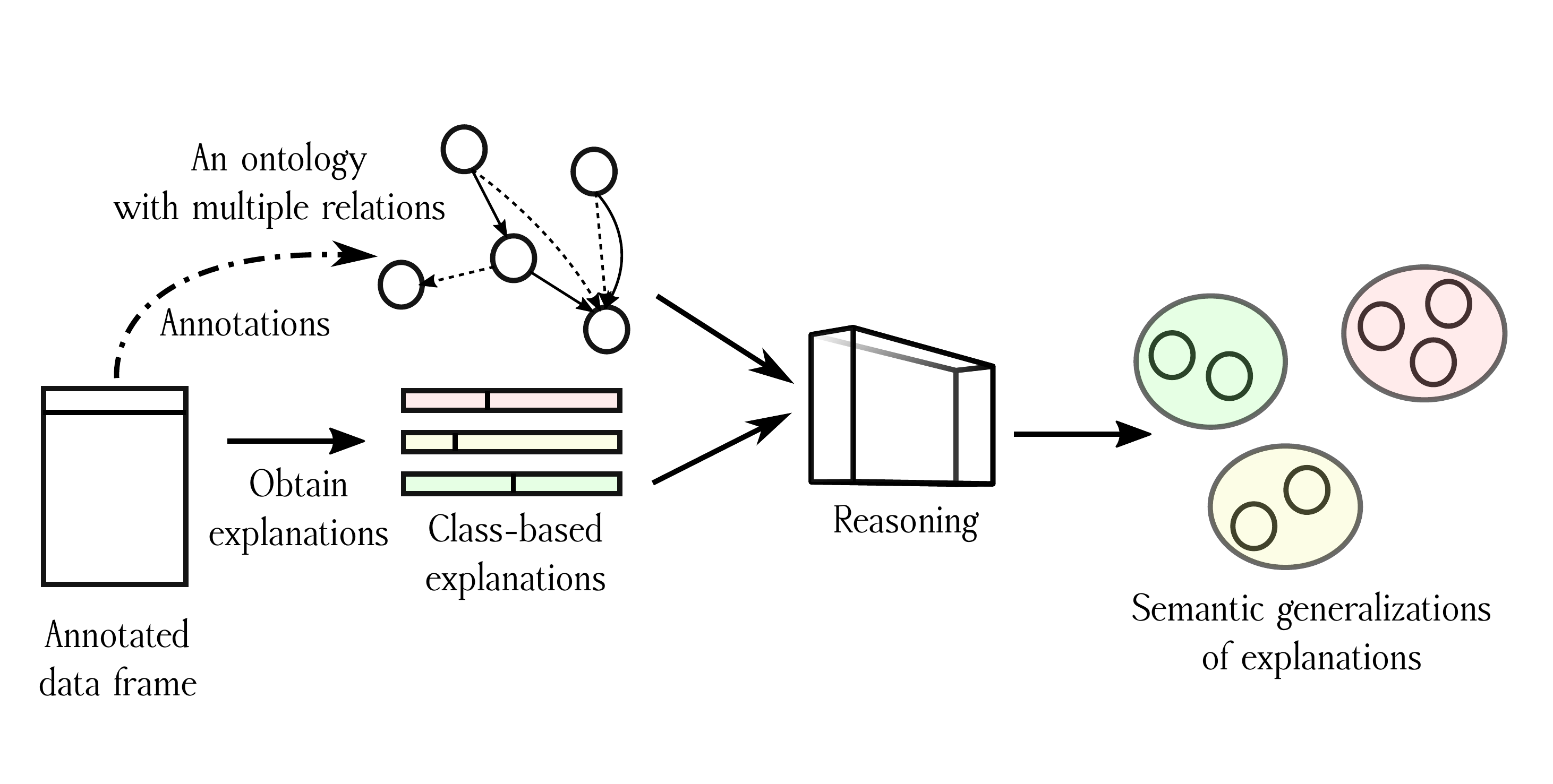}
    \caption{Overview of the proposed ReEx. Given a data set where the attributes map to a \emph{domain ontology}, ReEx first produces and aggregates the instance-level explanations. Top attributes are selected iteratively (vertical lines in the second sub-image of the three vectors), followed by a reasoning procedure, which exploits the explicitly given relations (within the ontology) to generalize the mapped terms into \emph{higher-level} semantic terms. Generalizations are obtained for each class (last sub-figure).}
    \label{fig:reasoning-scheme}
\end{figure}
The following section discusses the proposed ReEx approach,  summarized in Figure~\ref{fig:reasoning-scheme}.
The first part of ReEx concerns with obtaining class-specific aggregates of feature importances. To achieve this, the following steps are considered:

\textbf{Initial feature pruning.} We select the top $k$ features according to the descending order of their mutual information with the target variable. This myopic measure is selected due to its computational efficiency~\cite{mutualInfo}.

\textbf{Cross-validation and explanation acquisition.} The model is trained in a 10-fold cross-validation scheme, where for each of the test instances (one of the folds), the SHAP kernel explainer \cite{NIPS2017_7062} is used to obtain a list of Shapley values, indicating which features were the most relevant for a given prediction.

\textbf{Explanation aggregation.} The following procedure is considered for each of the classes; for each feature that appeared in an explanation at least once, the values across all explained instances are \emph{averaged}. When all features are considered, they are sorted in decreasing order based on the relevance values. 
 More formally, let $p_{C,+}$ denote a probability distribution of a collection of explanation values across the correctly classified samples of selected class $C$. Let $X_i$ denote the random variable representing the explanation for the $i$-th feature $F_i$. Let $\boldsymbol{X}$ denote the vector of all such random variables, having each entry associated with a given ($i$-th) feature. ReEx constructs a set of tuples defined as follows:
\begin{equation*}
\textsc{ReEx-agg} = \bigcup_C (C, \mathbb{E}_{\boldsymbol{X} \sim p_{C,+}} [\boldsymbol{X}]),
\end{equation*}
\noindent Here, $\boldsymbol{X}_i$ represents a random variable representing the Shapley value for the $i$-th feature. A vector of the expected values is thus considered for each class. The elements of the vector are the \emph{importances} associated with a given ($i$-th) feature.

A more detailed overview of the idea behind SHAP is discussed next.
SHAP \cite{NIPS2017_7062} is based on the \emph{coalitional game theory}, and aims to capture the importance of interactions between features via Shapley values.
When considered in a feature importance estimation scenario, the contribution of the $i$-th instance, denoted with $\tau_i$ is approximated by SHAP with the following expression:
\begin{equation*}
    \tau_i = \underbrace{\sum_{S \subseteq F \setminus \{i\}} \frac{|S|!(|F| - |S| - 1)!}{|F|!}}_{\textrm{All possible subsets}} 
    \underbrace{\bigg [f(x_{S \cup \{i\}}) - f(x_S)   \bigg ]}_{\textrm{Difference in predictive performance}}
\end{equation*}

\noindent where $S$ is a subset of all features $F$, $f$ is the used predictive model, and $x_S$ is an instance containing only features from the subset $S$.
Shapley values offer insights into the instance-level predictions by assigning fair credit to individual features for participation in interactions. They are commonly used to understand and debug black-box models \cite{ghosal2018explainable}. 

In this work, we use the SHAP \emph{kernel approximator}, the recently introduced, model-agnostic method for explaining model outputs.
The used SHAP kernel explainer is considered an additive feature attribution method. Such methods are characterized by having an explanation model $g$ that is a linear function of binary variables:
\begin{equation*}
g(z') = \phi_0 + \sum_{i = 1}^{|F|}\phi_i \cdot z_i'
\end{equation*}
\noindent where $z' \in \{0,1\}^{|F|}$, $|F|$ is the number of input features and $\phi_i \in \mathbb{R}$. This class of models assign an importance $\phi_i$ to each feature and, summing the effects of all such feature attributions, approximates the output $f(x)$ of the original model.
Detailed theoretical analysis of how this idea can be extended to an approximation of outputs via a kernel is given in \cite{NIPS2017_7062}. 
The final result of this step are thus lists of tuples for each class.

\textbf{Thresholding} - ReEx implements a \emph{dynamic threshold} for each class which we lower until at least $n$ number of features in that class have their assigned SHAP value greater than the threshold.

\textbf{Reasoning}. After the discussed procedure ReEx obtains a set of features that were estimated as the most important for each class. These collections are used as the input to reasoning, discussed next.
\emph{Selective staircase} is the faster of the two algorithms proposed as a part of this work. Given the threshold parameter, the algorithm generalizes a term into its highest ancestors, whose ratio of terms, connected to the other classes, is below the user-specified threshold. It searches through all viable ancestors in iterations, each iteration considering more general ancestors. Hence, the algorithm \emph{is deterministic}.
\emph{Ancestry} is a slower algorithm that considers \emph{lowest common ancestors} of a pair of terms. This means that every term that is the result of a step of generalization is connected to at least two previously found terms (or starting terms). Found ancestors are considered based on their ratio of connected starting terms from other classes and length of the path from the two generalized terms and their ancestor. The algorithm is non-deterministic since there is a random element involved when choosing a pair of terms used for finding their common ancestor.
It operates under the assumption, that generalization needs to be conducted based on \emph{pairs of terms}, rendering the method potentially more \emph{robust to noise}, however less flexible.

\section{Formulation of generalization approaches}
\label{sec:implementation}
In the following sections, we present the proposed implementations of the generalization procedures employed by ReEx. We begin with the Selective staircase (Section~\ref{sec:ss}), followed by the Ancestry algorithm (Section~\ref{sec:anc}).
\subsection{Selective staircase}
\label{sec:ss}
 In a single iteration, we create a set of parents of the terms in the set for each class based on a given domain ontology. These are \emph{more general}, however directly connected to the terms we currently have in the set. For each element in this set of parents, we calculate the proportion of starting terms of other classes that are descendants of this term. From this set, we remove those of which computed ratio is above the \emph{given threshold}.
We then add acceptable elements for each class from the parent set to the term set and remove from the term set the elements that are children of the newly added ones (we remove those that were generalized).
The step is repeated for each class until term sets stop changing (generalization stops).

\begin{algorithm}
\SetKwData{Left}{left}\SetKwData{This}{this}\SetKwData{Up}{up}
\SetKwFunction{Union}{Union}\SetKwFunction{FindCompress}{FindCompress}
\SetKwInOut{Input}{input}\SetKwInOut{Output}{output}
\Input{starting term sets, background knowledge, threshold}
\Output{a generalized term set for each class and depth of generalization for each term}
\While {not all converged}{
\For{$class \in $classes}{\label{forins}
$change \leftarrow $false\;
$parents\leftarrow $getParents({termSets[class]})\;
\For{$parent \in  $parents}{\label{forins2}
\If{intersectionRatio$(parent) <= $threshold}{
termSets[class].add(parent)\;
termSets[class].removeChild(parent)\;
$change \leftarrow $true\;}}
\If{not change}{
converged[class]$\leftarrow$ true\;}}}
\caption{Selective staircase - generalizes every term into all of its ancestors, then selects those suitable.}
\end{algorithm}
\subsection{Ancestry}
\label{sec:anc}
For each term in the term set of a given class, we select another term in the same term set and find their \emph{lowest common ancestor}. Note that the depth of this generalization step is considered to be equal to $\min$(distance(first term, ancestor), distance(second term, ancestor)) -- \emph{the shortest path to the ancestor}. We calculate the ratio of the starting terms of other classes that are descendants of this ancestor. If the following inequality: 
$\textrm{ratio} / (\textrm{depth} \cdot \textrm{weight}) - \frac{1}{2} < 0$
holds for a given weight (parameter), the ancestor is added to the generalized term set of the class.
After repeating this step with the whole term set, we remove the terms that were used in finding an ancestor that was added to the term set.
We iterate this step for each class until the term sets stop changing (generalization stops).
\begin{algorithm}
\SetKwData{Left}{left}\SetKwData{This}{this}\SetKwData{Up}{up}
\SetKwFunction{Union}{Union}\SetKwFunction{FindCompress}{FindCompress}
\SetKwInOut{Input}{input}\SetKwInOut{Output}{output}
\Input{starting term sets, background knowledge, weight}
\Output{a generalized term set for each class and depth of generalization for each term}
\While {not all converged}{
\For{$class \in $classes}{\label{forins3}
$change \leftarrow $false\;
\For{$term \in  $termSets[class]}{\label{forins4}
$randomTerm \leftarrow selectRandomTerm(termSets[class])$\;
$(ancestor, depth) \leftarrow findAncestor($term, randomTerm)\;
\If{intersectionRatio($ancestor) / (depth * weight) < $0.5}{
termSets[class].add(ancestor)\;
setUsed(term, randomTerm)\;
$change \leftarrow $true\;}}
removeUsed(termSets[class])\;
\If{not change}{
converged[class]$\leftarrow$ true\;}}}
\caption{Ancestry - generalizes pairs of terms into their lowest common ancestor, then checks whether the ancestor is suitable.}
\end{algorithm}

\subsection{Measuring success}
\label{genq}
Information content can be defined for a given term $t$ as:
$\textsc{IC}(t) = -\log ( p(t) )$,
\noindent where $p(t)$ represents the prior probability of a given entity being annotated with the term $t$. 
The information content offers direct insight into how general or specific a particular term or a set of terms is (e.g., mean IC). 
Let $\mathfrak{O}$ represent the set of all terms in a given ontology and $T$ the set of considered terms.
As the point of this paper is to assess the generalization, we propose the following measure, offering direct insight into the generalization quality,
\begin{equation*}
    \begin{aligned}
    \textsc{GenQ}(T)= 1 - \frac{\sum_{t \in T} \log(p(t))}{|T| \cdot \textsc{NrO}}\quad \\ \textrm{where } \textsc{NrO} = -\max(IC({t \in \mathfrak{O}})) \sim \log (1/|\mathfrak{O}|).
    \end{aligned}
\end{equation*}
\noindent The $\textsc{GenQ(T)}$'s domain is between zero and one. Intuitively this score can be understood as one minus the normalized average information content. The normalization is ontology-specific, i.e., prior to computing the score, ReEx is capable of identifying the term with the maximum information content -- such terms are \emph{commonly} the ones that appear only once (hence the term $\log (1/|\mathfrak{O}|)$).

\section{A toy example and some theoretical properties}
\label{sec:example}
\begin{figure}[ht]
\centering
\captionsetup{width=\linewidth}
\begin{tabular}{ccc}
\subcaptionbox{Starting terms}{\includegraphics[width = 1in]{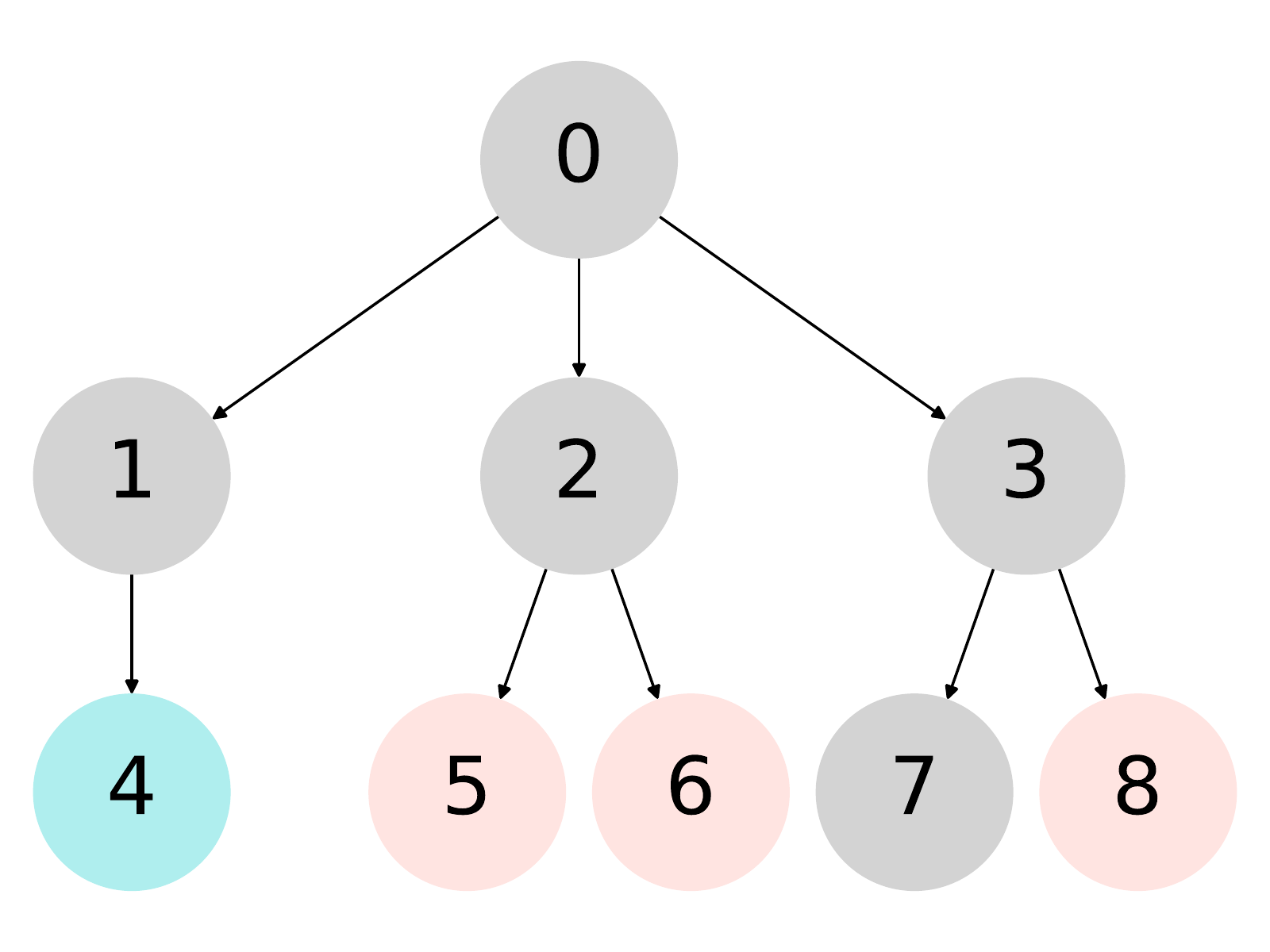}} &
\subcaptionbox{Selective\\ staircase}{\includegraphics[width = 1in]{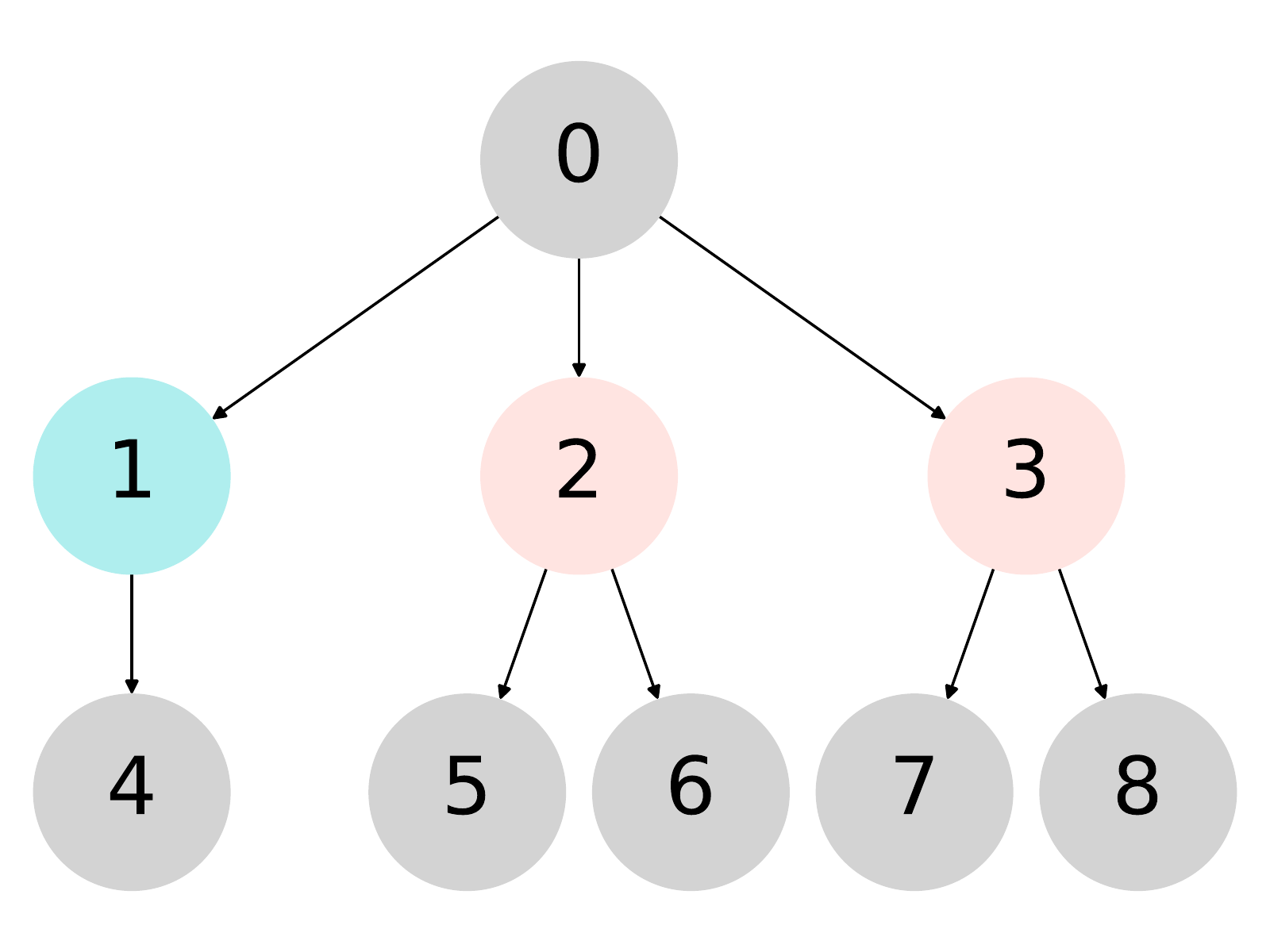}} &
\subcaptionbox{Ancestry}{\includegraphics[width = 1in]{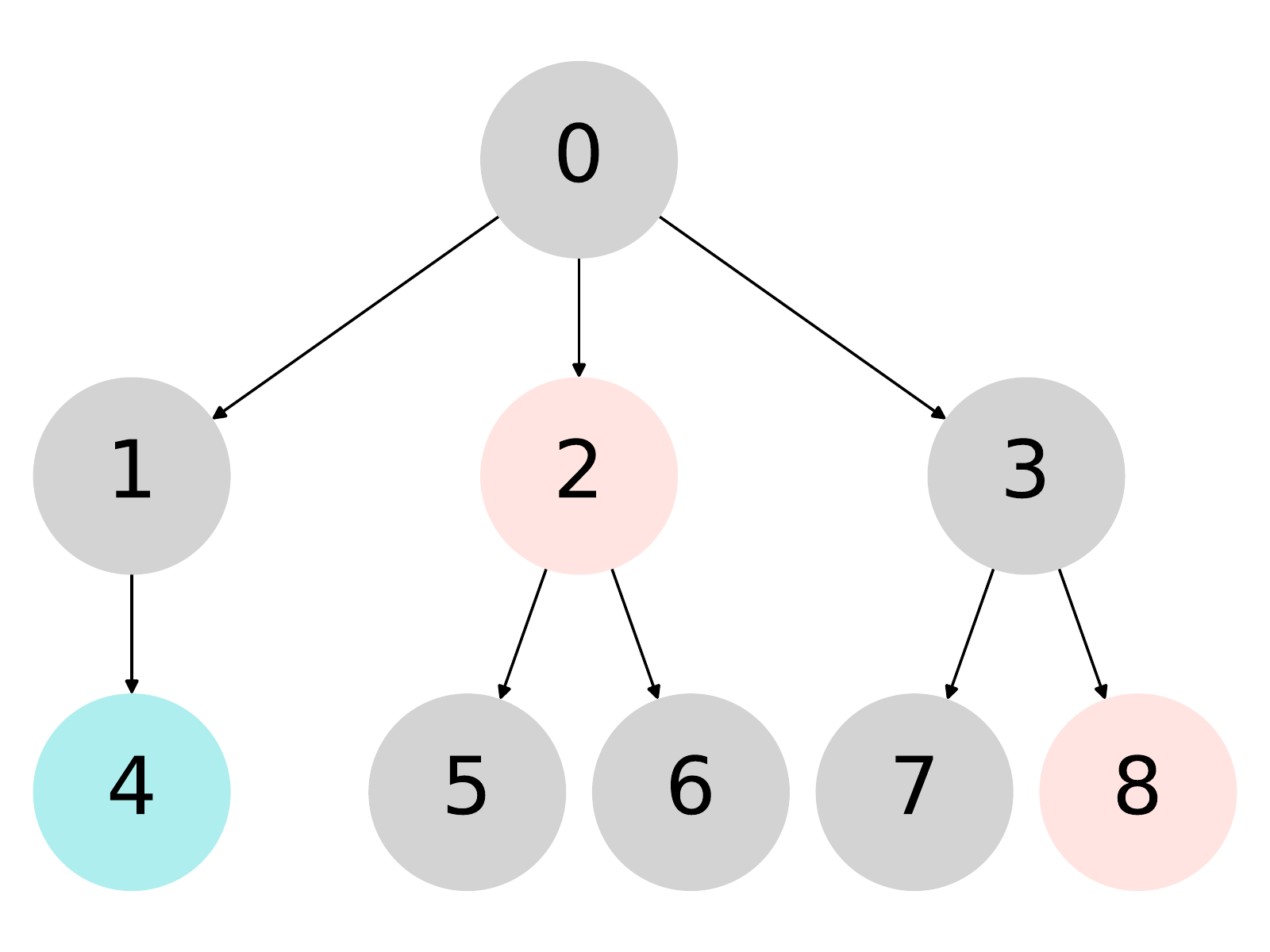}} \\
\end{tabular}
\caption{A toy example of the two generalization procedures considered. The red and green colors represent the terms belonging to a particular class. They are generalized to the point no more generalization is possible without the term co-occurring as a result for both classes.}
\label{fig:bfg}
\end{figure}

This section presents the main idea of the two algorithms in the form of a simple toy example when performing generalization on directed acyclic graphs (the example includes trees).
Our example includes two classes - red and green and their sets of starting terms as indicated in figure Figure~\ref{fig:bfg}.
The set of starting terms for the green class is \{4\} and for the red class it is \{5, 6, 8\}.
We run the Selective staircase on our example with the threshold parameter of 0, meaning that terms that are the result of the generalization must not have any starting terms of other classes as their descendants.
In the first iteration, term 4 will get generalized into term 1, terms 5 and 6 into term 2 and term 8 into term 3. After the first iteration sets of terms will therefore be \{1\} (for the green class) and \{2, 3\} (for the red class).
In the second iteration, all terms in term sets have the same parent - 0, but because 0 is connected to starting terms of both classes, it cannot be added in either term set. Because term sets stay the same in the second iteration, generalization stops.
We run Ancestry on our example with the weight parameter set to 0.000001, which, for our example, effectively means, that terms which are the result of the generalization must not have any starting terms of other classes as their descendants.
Results of generalization may vary because the algorithm is \emph{not deterministic}, so we describe the steps that the algorithm can make.
Green class' term set holds one term and will not be generalized, since two terms are necessary to find their lowest common ancestor. If the algorithm tries to generalize 8 with 5 or 6, their \emph{lowest common ancestor} is 0, which has a descendant of another class, but if it generalizes 5 and 6, the result of the generalization will be 2.
If 5 and 6 are not generalized into 2 in the first iteration, generalization will stop, since it hasn't made any progress in the iteration. If they are, however, we proceed to the second iteration in which 0 is found as the ancestor of 2 and 8, but is not acceptable, so the generalization stops.

\section{Empirical evaluation}
\label{sec:empirical}
In the following section we discuss the conducted experiments, aimed at clarifying the capabilities of the two considered generalization schemes across a wide array of real-life data sets.
The data sets considered are described in Table~\ref{tab:datasets}. The considered selection includes data sets where the task is phenotype prediction (e.g., tissue classification).
\begin{table}[b!]

\caption{Statistical Properties of the Considered Data Sets.}
\label{tab:datasets}
    \centering
    \resizebox{.4\textwidth}{!}{
\begin{tabular}{lrrrr}
\toprule
  Dataset &  Instances &  Features &  Classes \\
\midrule
  DLBCL B \cite{tissues} &        180 &       661 &        3  \\
  DLBCL A \cite{tissues} &        141 &       661 &        3  \\
 Breast A \cite{tissues} &         98 &      1213 &        3  \\
 Breast B \cite{tissues} &         49 &      1213 &        4 \\
  DLBCL D \cite{tissues} &        129 &      3795 &        4  \\
  DLBCL C \cite{tissues} &         58 &      3795 &        4  \\
  Multi A \cite{tissues} &        103 &      5565 &        4 \\
  Multi B \cite{tissues} &         32 &      5565 &        4  \\
     TCGA \cite{weinstein2013cancer} &        801 &     20531 &        5  \\
\bottomrule
\end{tabular}}
\end{table}
The used Gene Ontology \cite{gene2019gene} consists of 44{,}700 nodes and 91{,}526 edges. Considered relations are "part-of", "regulates", "negatively-regulates", "positively-regulates" and "is-a"\footnote{All relations are intentionally considered, to explore ReEx's capability to operate in automated manner -- without human interventinons.}.

Finally, the \emph{Mapping} from genes to corresponding Gene Ontology terms consists of 19{,}412 genes mapped on average to 14.82 GO terms.
The task considered in this work is \emph{multiclass classification}. The methods used during evaluation are specified as follows.
We employ three different classifiers, namely Gradient Boosting Machines (gradient-boosting), Random Forest (random-forest) and Support Vector Machine classifiers (SVM). Each of the classifiers is explained with \cite{NIPS2017_7062}; the explanations further used as discussed in Section~\ref{sec:reex}. For both considered reasoning algorithms (Ancestry and Selective staircase), we computed the following grid of configurations and reported the mean with the standard deviation; subset sizes were either 100 or 5000 (initial pruning), absolute Shapley values were considered, threshold parameter, when using Selective staircase, was either 0,0.2 or 0.4 and weight parameter, when using Ancestry, was either 0.000001, 0.3, 0.6 or 3. The minimum number of terms was set to 10 (terms used for reasoning) and the thresholding step size was set to 0.975. Hence, around 500 different configurations were computed, when evaluating all the mentioned settings on all nine data sets.
As the purpose of this work is to demonstrate that simple and efficient generalization is possible, we compare the algorithms' performances against the baseline we define as the na\"ive, direct mapping of features to the collection of terms -- the mapping defined as part of the Gene Ontology. The rationale for introducing this baseline is as follows. Should the \textsc{GenQ} improve upon the na\"ive generalization (mapping), it will be higher. On the contrary, \textsc{GenQ} that would be lower than the generic mapping could indicate the algorithm was in fact performing
\emph{specialization}. We next present the obtained results alongside their discussion.

\section{Results -- reasoning behavior}
\label{sec:results}
\begin{figure}[h!]
\centering
\captionsetup{width=\linewidth}
\begin{tabular}{cc}
\subcaptionbox{Generalization performance w.r.t. different learners. \label{graphA}}{\includegraphics[width = .45\linewidth]{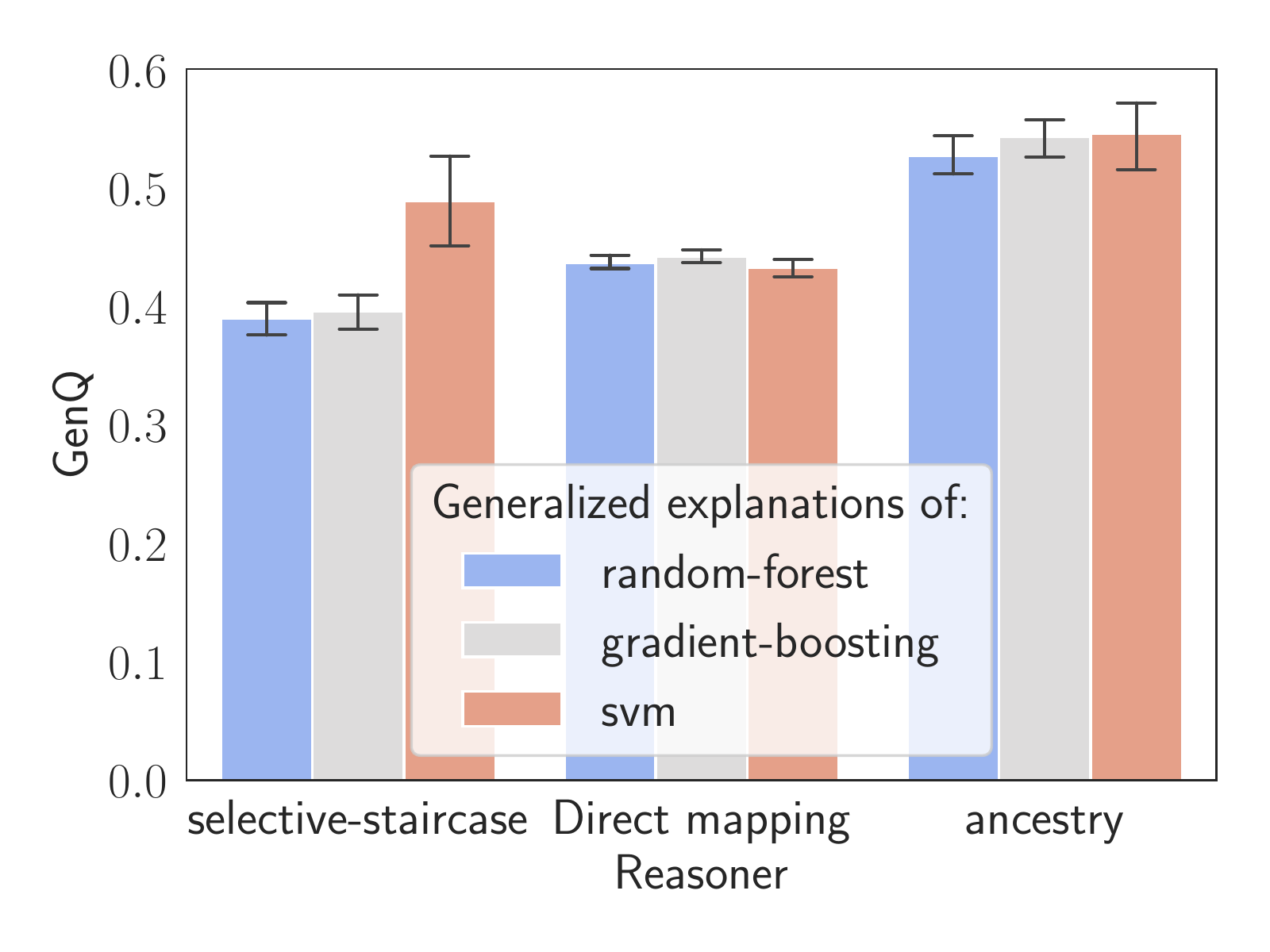}}
&
\subcaptionbox{Number of terms in the final generalization.\label{graphB}}{\includegraphics[width = .45\linewidth]{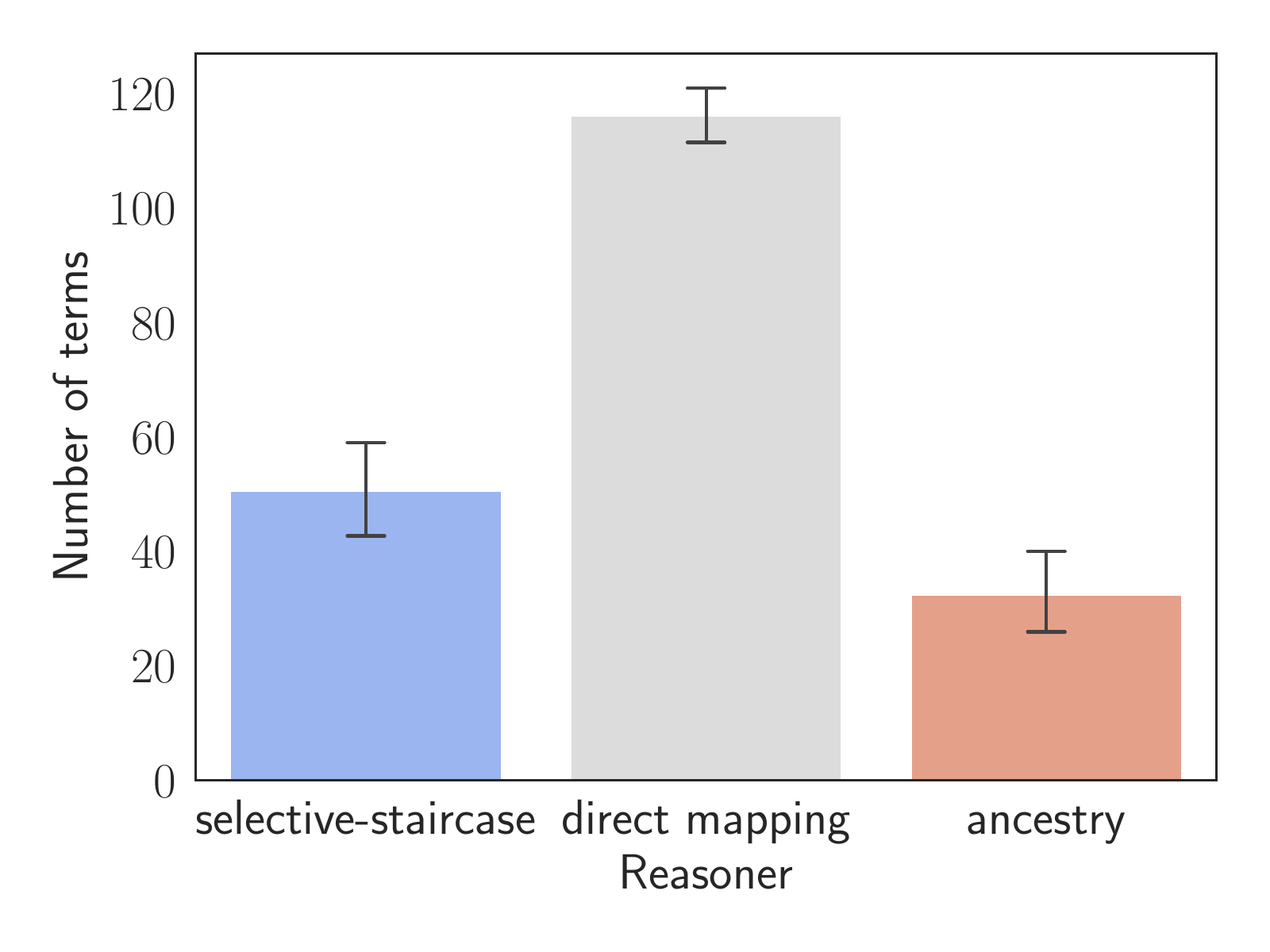}}
\\
\subcaptionbox{Generalization depth w.r.t. threshold parameter - Selective staircase algorithm.\label{graphC}}{\includegraphics[width = .45\linewidth]{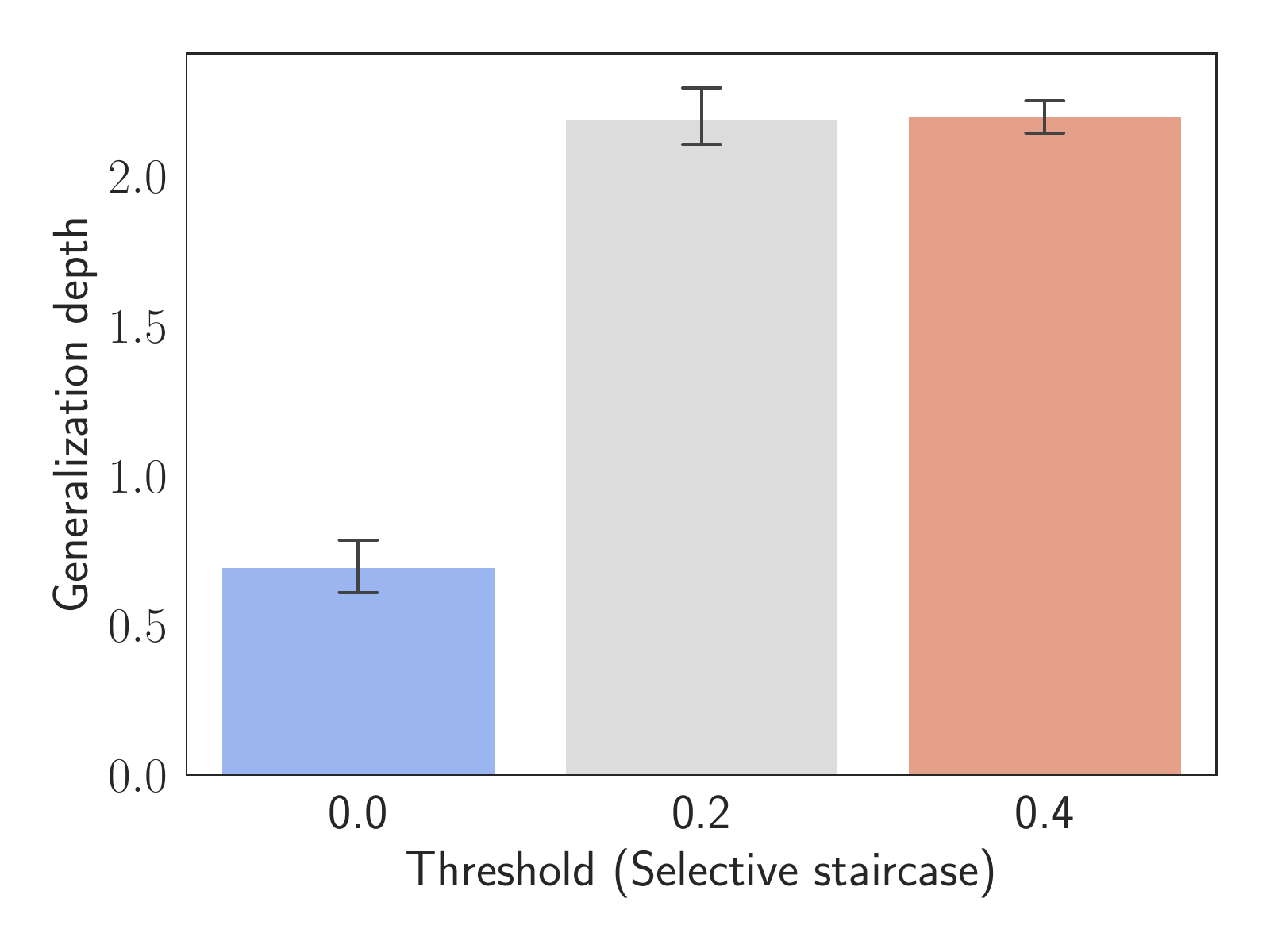}}
&
\subcaptionbox{Generalization performance w.r.t. threshold parameter - Selective staircase algorithm.\label{graphD}}{\includegraphics[width = .45\linewidth]{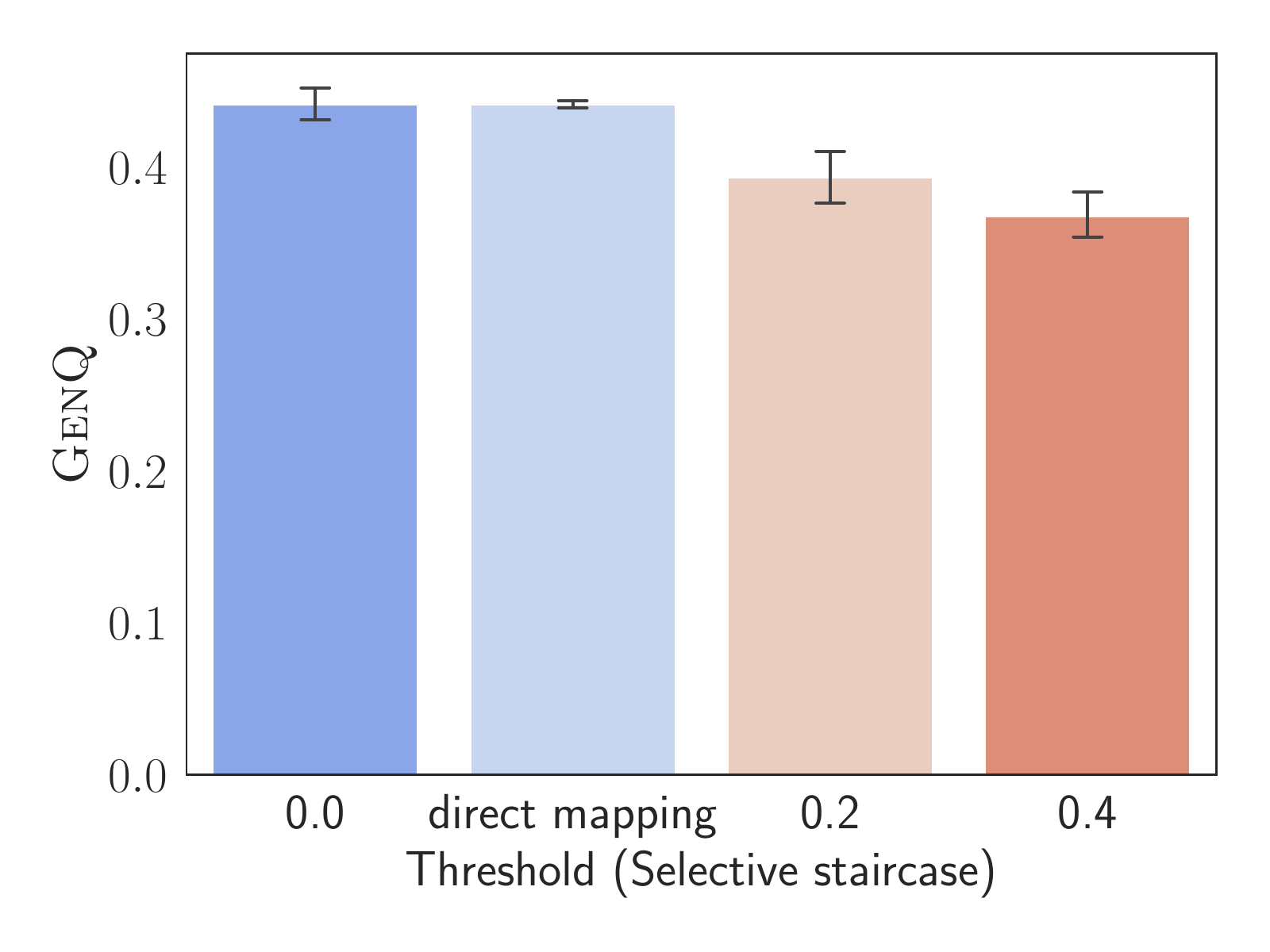}}
\\
\subcaptionbox{Generalization depth w.r.t. weight parameter - Ancestry algorithm.\label{graphE}}{\includegraphics[width = .45\linewidth]{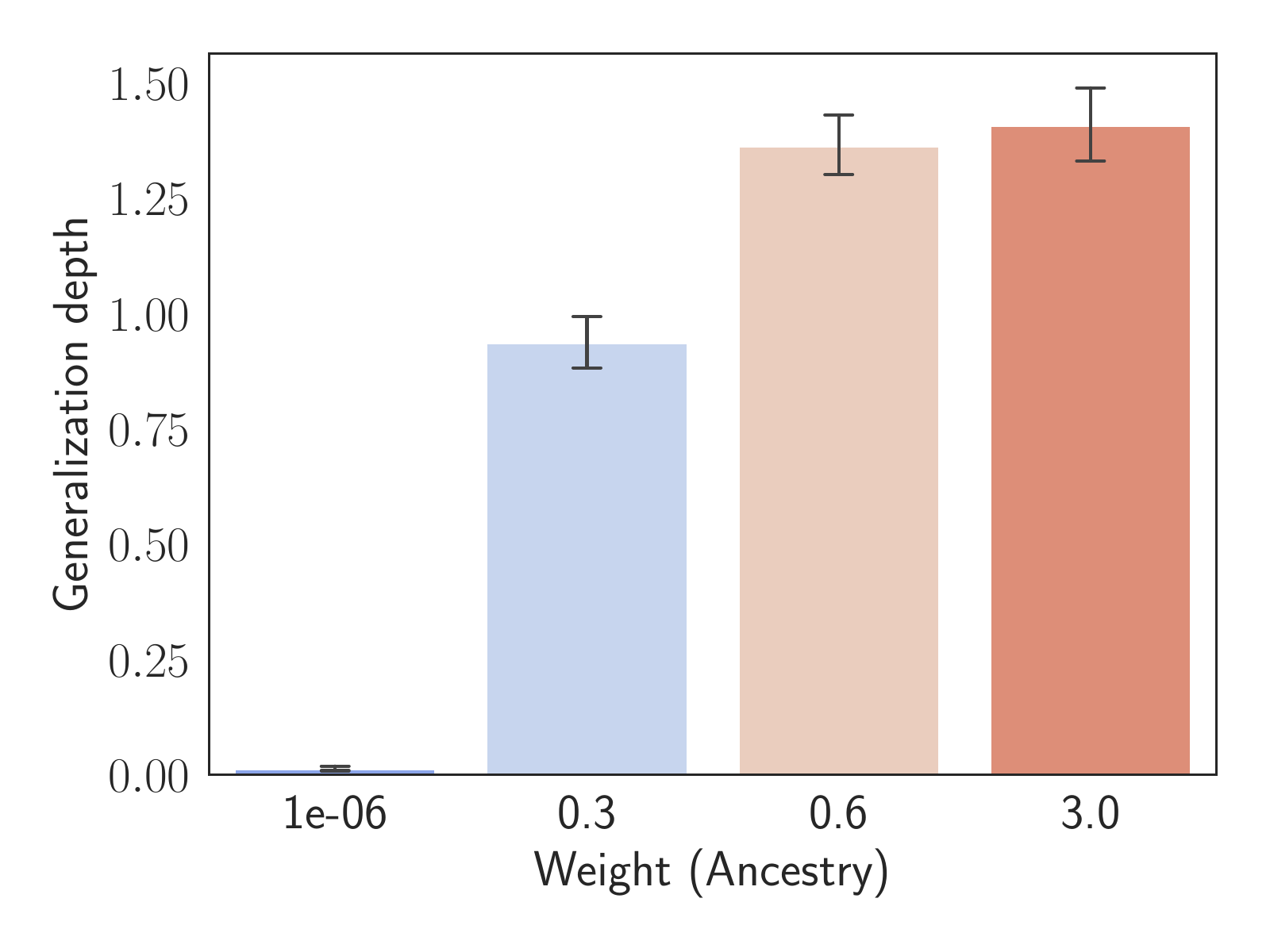}}
&
\subcaptionbox{Generalization performance w.r.t. weight parameter - Ancestry algorithm.\label{graphF}}{\includegraphics[width = .45\linewidth]{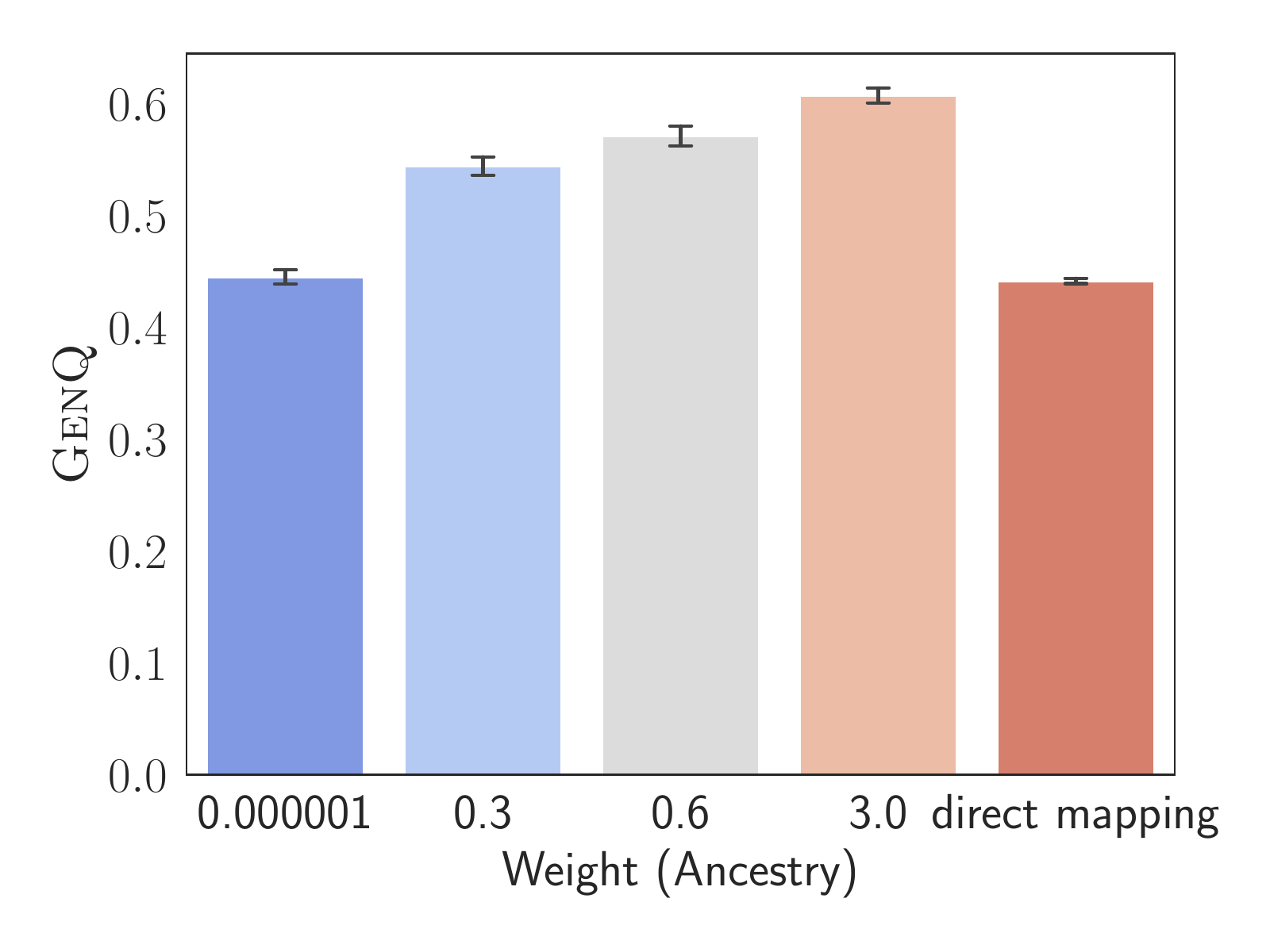}}
\\
\end{tabular}
\caption{Summary of the results across multiple hyperparameter settings.}
\label{fig:results-main}
\end{figure}

The following section includes the main results, obtained as the aggregate across the parameter space discussed in the previous section. The main results are summarized in Figure~\ref{fig:results-main}.
 It can be seen that the Ancestry method consistently generalizes terms into terms with a higher GenQ rating, while the Selective staircase's performance is less consistent \ref{graphA}. Both methods notably decrease the number of resulting terms \ref{graphB}. Parameters "threshold" and "weight" serve as the strictness of no cross-section between classes for Selective staircase and Ancestry respectively; smaller value meaning a more constrained generalization. Therefore the depth of generalization increases with increasing "threshold" and "weight" parameters, as seen in \ref{graphC} and \ref{graphE}. From graph \ref{graphD} we can see, that a less constrained generalization with the Selective staircase can decrease generalization performance, even though we achieve a higher generalization depth. Generalization performance of the Ancestry algorithm increases with a less constrained generalization, however \ref{graphF}. That could be due to Ancestry's generalization already being constrained with searching for ancestors of pairs of terms.
The code to reproduce the experiments is available to the reader\footnote{The official URL is: \url{https://github.com/OpaqueRelease/ReEx}}.

\section{Results -- Examples of semantic explanations}
Most general terms associated with classes are listed as follows\footnote{The star-marked entries' full names are DNA-binding transcription activator activity, RNA polymerase II-specific and detection of chemical stimulus involved in sensory perception of smell, respectively.} (Figure~\ref{fig:exp}).
\begin{figure}[h!]
\vspace{-0.6cm}
\begin{align*}
\textbf{Subtype 1} &:- \textrm{protein homodimerization activity} \\&\wedge \textrm{protein heterodimerization activity} \\ &\wedge \textrm{ubiquitin protein ligase binding}\\ &\wedge \textrm{magnesium ion binding}\\ &\wedge \textrm{calmodulin binding}\\
\textbf{Subtype 2} &:- \textrm{ATP binding} \\&\wedge \textrm{protein homodimerization activity} \\&\wedge \textrm{neutrophil degranulation}\\&\wedge \textrm{DNA-binding transcription activator activity}^*\\&\wedge \textrm{sensory perception of smell}^*\\
\end{align*}
\vspace{-0.8cm}
\caption{Example of generalized explanations for the Breast A \cite{tissues} data set (classes are different subtypes).}
\label{fig:exp}
\end{figure}

The terms comprising an explanation are human-understandable and as such offer direct insight into the biological processes governing the space of instances belonging to a given class.
A single process can be attributed to more than one class, depending on how constrained the generalization has been.
The two key aspects of the obtained \emph{logical explanations} are: First, the logical statements for individual classes (conjuncts of semantic terms) are notably different -- indicating different biological processes underlying the successful classification into a given class. And second, the explanations are directly understandable and can be verified by a domain expert within seconds. On the contrary, should raw feature names be used (gene names in this case), if the domain expert does not know all the names, as well as the associated processes by heart, the expert is expected to perform time-demanding manual literature search.

\section{Conclusions}
\label{sec:conclusions}
In this work, we proposed ReEx, one of the first approaches capable of semantic generalization of model explanations obtained by contemporary tools such as SHAP. The work evaluates two different reasoning paradigms (Selective staircase and Ancestry), showing both schemes out-perform generic generalization commonly employed by e.g., statistical enrichment analysis approaches. Further, we demonstrate how ReEx can produce logical explanations comprised of semantic term conjuncts, specific for individual classes -- these types of explanations offer direct insight into the e.g., biological background relevant to classifying a given instance into a particular class.
Understanding how biological context can be exploited for obtaining more interpretable explanations could also be extended to contemporary knowledge graphs, where the data is semi-automatically curated. Here, the amount of noise is potentially larger than when considering e.g., the Gene Ontology (this work), which we believe is an issue to be addressed in future work.
The proposed work was implemented in the form of a simple-to-use Python library, compatible with existing machine learning pipelines. Albeit being one of the first studies to explore the potentials of semantic generalization of black-box model explanations, ReEx could be further analyzed and improved.

\section{Acknowledgements}
The work of the last author was funded via a young researcher grant (ARRS). The work of other authors was supported by the Slovenian Research Agency (ARRS) core research programs P2-0103 and P6-0411, 
and research projects J7-7303, L7-8269, and N2-0078 (financed under the ERC Complementary Scheme). The work was also supported by European Union's Horizon 2020 research and  innovation programme under grant agreement No 825153, project EMBEDDIA (Cross-Lingual Embeddings for
Less-Represented Languages in European News Media).
\bibliographystyle{IEEEtran}
\bibliography{refs}

\end{document}